\documentclass[pmlr,twocolumn,10pt]{jmlr} 
\usepackage{booktabs}
\usepackage{siunitx}
\usepackage[switch]{lineno}

\usepackage{multirow}
\usepackage{array}
\usepackage{newtxtext,newtxmath} 

\theorembodyfont{\upshape}
\theoremheaderfont{\scshape}
\theorempostheader{:}
\theoremsep{\newline}

\jmlrvolume{LEAVE UNSET}
\jmlryear{2026}
\jmlrsubmitted{LEAVE UNSET}
\jmlrpublished{LEAVE UNSET}
\jmlrworkshop{Conference on Health, Inference, and Learning (CHIL) 2026} %
\title[Short Title]{From Wearables to Warnings: Predicting Pain Spikes in Patients with Opioid Use Disorder}

\author{%
\Name{Abhay Goyal}
\Email{abhay@nimblemind.ai} \\
\addr Nimblemind, USA
\AND
\Name{Navin Kumar} \Email{navin@nimblemind.ai}\\
\addr Nimblemind, USA
\AND
\Name{Kimberly DiMeola}
\Email {kdimeola@aptfoundation.org} \\
\addr APT Foundation, New Haven, CT, USA
\AND
\Name{Rafael Trujillo}\\
\addr Florida International University, Miami, FL, USA
\AND
\Name{Soorya Ram Shimgekar} \Email{sooryas2@illinois.edu}\\
\addr University of Illinois -- Urbana Champaign
\AND
\Name{Christian Poellabauer}
\Email{cpoellab@fiu.edu}\\
\addr Florida International University, Miami, FL, USA
\AND
\Name{Pi Zonooz} \Email{pi@nimblemind.ai}\\
\addr Nimblemind, USA
\AND
\Name{Ermonda Gjoni-Markaj}\\
\addr APT Foundation, New Haven, CT, USA
\AND
\Name{Declan Barry}
\Email{declan.barry@yale.edu}\\
\addr APT Foundation, Yale University School of Medicine, New Haven, CT, USA
\AND
\Name{Lynn Madden}
\Email
{lmadden@aptfoundation.org} \\
\addr APT Foundation, Yale University School of Medicine, New Haven, CT, USA
}

\begin{document}

\maketitle

\begin{abstract}
Chronic pain (CP) and opioid use disorder (OUD) are common and interrelated chronic medical conditions. Currently, there is a paucity of evidence-based integrated treatments for CP and OUD among individuals receiving medication for opioid use disorder (MOUD). Wearable devices have the potential to monitor complex patient information and inform treatment development for persons with OUD and CP, including pain variability (e.g., exacerbations of pain or pain spikes) and clinical correlates (e.g., perceived stress). However, the application of large language models (LLMs) with wearable data for understanding pain spikes, remains unexplored. Consequently, the aim of this pilot study was to examine the clinical correlates of pain spikes using a range of AI approaches. We found that machine learning models achieved relatively high accuracy ($>$0.7) in predicting pain spikes, while LLMs were limited in providing insights on pain spikes. Real-time monitoring through wearable devices, combined with advanced AI models, could facilitate early detection of pain spikes and support personalized interventions that may help mitigate the risk of opioid relapse, improve adherence to MOUD, and enhance the integration of CP and OUD care. Given overall limited LLM performance, these findings highlight the need to develop LLMs which can provide actionable insights in the OUD/CP context.
\end{abstract}

\paragraph*{Data and Code Availability}
Data and code is available upon reasonable request. We collected data from 25 patients with chronic pain and opioid use disorder from the APT Foundation. 
\paragraph*{Institutional Review Board (IRB)}
The study was approved by the Yale University Institutional Review Board (2000034714) and the APT Foundation Board of Directors.

\section{Introduction}
\label{sec:intro}
Chronic pain (CP) and opioid use disorder (OUD) are common and interrelated chronic medical conditions. Chronic pain is defined as noncancer pain lasting most days for at least three months. The Centers for Disease Control and Prevention (CDC) estimates that 20\% of adult Americans experience at least one lifetime episode of chronic pain and 8\% have high-impact chronic pain that limits their life or work activities on most days
\cite{dahlhamer2016helmick}. 
Over two million individuals in the United States have OUD, which is a chronic relapsing condition that is associated with high rates of mortality
\cite{ahmad2023provisional}. 
Across the 12-month period ending in December 2024, the Centers for Disease Control and Prevention (CDC) reported an estimated 80,391 drug overdose deaths in the United States
\cite{CDC_NCHS_2025_provisional2024}. 

Estimates suggest that at most 20\% of adults with OUD are on MOUD, and 2019 adjusted estimates suggest past-year OUD affected over seven million people in the US \cite{krawczyk2022has}. MOUD has shown several benefits such as decreases in mortality, increases in treatment adherence, decreases in heroin use, and augmented health, social and criminal justice outcomes \cite{organek2022bitter,kumar2021role,mattick2009methadone,mun2019pain}. Co-management of opioid use disorder and chronic pain is a growing clinical challenge: Prevalence estimates of CP among patients on opioid agonist treatment (e.g., methadone or buprenorphine) are higher than the general population (37-61\% vs. 20\%) and may reflect reduced pain thresholds, increased pain sensitivity, or opioid-induced hyperalgesia \cite{su2025addressing}. Treatment of patients with CP and OUD presents challenges as unrelieved pain is associated with illicit opioid and alcohol use, anxiety, and depression \cite{su2025addressing}, and OUD may in some cases develop as a result of opioid therapy for CP. 

Among patients with OUD, CP is associated with opioid cravings and pain exacerbation or spikes are linked to opioid use. However, chronic pain is not routinely assessed or addressed in opioid treatment programs, nor is OUD routinely assessed in pain management programs. Currently, there is a paucity of evidence-based integrated treatments for CP and OUD among individuals receiving MOUD \cite{mun2019pain}. A greater understanding of pain variability (including pain spikes) and clinical correlates (e.g., perceived stress) among patients receiving MOUD with CP may inform the development of an integrated treatment intervention targeting CP and OUD. We defined pain spikes as experiencing $>70$th percentile of self-reported pain. 

Recent work has explored the utility of collecting smartwatch sensor data among patients in outpatient MOUD treatment and of examining the utility of digital sensing in predicting MOUD treatment retention, opioid use events, and medication adherence \cite{mahoney2023identifying} . Such devices harness a patient-friendly automated sensing system to measure individual outcomes, such as sleep, vital signs, and physical activity, to inform treatment \cite{hsaio2024associations,martinez2020associations}. Thus, wearable devices have the potential to monitor complex patient information and provide information to inform treatment development for persons with OUD and CP. Recent advancements in artificial intelligence (AI), particularly large language models (LLMs), have demonstrated the ability to extract key information from electronic health records (EHRs) and wearable sensors, offering predictive capabilities for both substance use and pain-related outcomes \cite{gabriel2023leveraging,mahbub2024leveraging}. However, the application of LLMs in integrating wearable and EHR data for understanding dynamic clinical events, such as pain spikes, remains underexplored. Our study seeks to address this gap by evaluating how multiple LLMs interpret and analyze data (survey, wearable, medication data) from patients with OUD and chronic pain. We outline a proof-of-concept study that used wearable devices to obtain longitudinal data from patients with OUD and CP, to provide data that may eventually lead to more individualized or tailored treatment provision. The overall aim of this pilot study was to examine the clinical correlates of pain spikes using a range of AI approaches.

\section{Related Work}
\textbf{Opioid use and wearable devices}
The integration of wearable technology into opioid use disorder (OUD) research has expanded rapidly in recent years. Multiple studies have explored how such devices can support the detection, monitoring, and prevention of opioid misuse and overdose. For instance, during the COVID-19 pandemic, a survey examined attitudes toward overdose-detection wearables among individuals with prior opioid use \cite{kanter2021willingness}. Most participants indicated that they would be willing to wear a device capable of automatically identifying an overdose and delivering a reversal agent. Respondents emphasized comfort and discretion as critical factors influencing long-term adoption.

Other work has focused on the technical development of biosensor systems. One team proposed a wrist-worn platform that analyzed electrodermal activity, skin temperature, and motion data to recognize opioid intake in real time with 99\% accuracy \cite{mahmud2018automatic}. The framework was adaptable for both clinical and at-home monitoring and showed promise for identifying other substances as well. A feasibility study with 16 participants diagnosed with OUD demonstrated that physiological indicators—such as respiratory rate and movement—captured by a wearable device could reliably reflect opioid use episodes \cite{roth2021wearable}.

Machine learning has further enhanced the analytical power of wearables in OUD research. Using biosensor data, \cite{kulman2021identifying} classified opioid withdrawal states with a Random Forest model (AUC = 0.9997). Similarly, wearables have been used to track post-surgical opioid self-administration, achieving validation accuracies above 80\% \cite{garcia2022using}. Accelerometry worn at the sternum has also been utilized to quantify restlessness related to withdrawal, showing strong associations between movement intensity and symptom severity \cite{lambert2022leveraging}.

Beyond monitoring, wearable devices are being integrated into behavioral interventions. A Stage 1 randomized controlled trial assessed a Zoom-based Mindfulness-Oriented Recovery Enhancement (MORE) program augmented with Just-in-Time Adaptive Intervention (JITAI) prompts driven by wearable sensor data \cite{garland2023zoom}. Participants receiving MORE + JITAI showed larger reductions in craving, stress, and pain and reported higher positive affect relative to those in a supportive psychotherapy control.

Collectively, these studies indicate that: (1) individuals with OUD are receptive to wearable overdose-detection tools, and (2) sensor-based systems, often paired with machine learning, can effectively detect opioid intake, withdrawal, and related behavioral changes in real time.
 
\textbf{Chronic pain and wearable devices}

Wearables have also been investigated for chronic pain (CP) management, providing both objective metrics and therapeutic feedback. In one home-based study, a movement-to-sound wearable helped participants regulate activity levels, fostering a sense of agency and supporting pacing strategies for pain management \cite{singh2017supporting}. Another study tested the feasibility of a sleep-monitoring EEG headband (Dreem 2) in adults with chronic pain; participants reported high comfort, usability, and willingness for long-term use \cite{zambelli2022exploring}.

A year-long clinical evaluation found that chronic pain patients using wearable health technology (WHT) experienced reductions in depressive symptoms and prescribed morphine milligram equivalents compared to control groups, suggesting benefits for both psychological well-being and medication management \cite{han2022long}. Other work demonstrated that nocturnal heart rate patterns could predict next-day pain intensity, implying the value of cardiac biomarkers for dynamic pain assessment \cite{dudarev2022night}.

More recently, digital biomarkers derived from wearables have been applied in spinal cord stimulation (SCS) therapy to model pain trajectories. By analyzing activity and physiological data, machine-learning algorithms successfully predicted pain intensity and patient-reported outcomes, supporting the feasibility of objective, data-driven treatment monitoring \cite{patterson2023objective}.

Together, this body of work shows that: (1) wearable devices can predict pain fluctuations and therapeutic responses, and (2) continuous sensor data can complement self-reported measures in chronic pain management. Yet, few studies have examined patients who simultaneously experience chronic pain and are receiving medications for opioid use disorder (MOUD). Moreover, research has rarely analyzed wearable-based correlates of acute pain spikes in this dual-diagnosis population. Addressing this gap, our pilot study explores clinical predictors of pain spikes among 20 individuals on MOUD with co-occurring chronic pain.

\textbf{LLMs, opioid use disorder, and chronic pain}
Recent advances in large language models (LLMs) have introduced new possibilities for analyzing complex clinical data in OUD and chronic pain research. Within perioperative care, transformer-based LLMs have been used to extract persistent postoperative opioid use information from electronic health records (EHRs), improving accuracy in identifying opioid-related outcomes \cite{gabriel2023leveraging}. Similarly, the Diet-ODIN framework combined graph-based learning with LLMs to investigate how dietary patterns influence opioid misuse, achieving strong predictive performance \cite{zhang2024diet}.

LLMs have also enhanced clinical data extraction: models such as Flan-T5 outperformed traditional rule-based methods in capturing OUD severity and diagnostic nuances from clinical text, aiding treatment planning and risk stratification \cite{mahbub2024leveraging}. In the context of chronic pain, LLMs have been applied to predict placebo responses in lower-back pain by identifying linguistic and psychological markers like hope and anxiety \cite{ap2025advancing}. GPT-4 has shown expert-level agreement when assessing pain narratives from fibromyalgia patients, suggesting potential for automating patient-reported assessments \cite{amidei2024can}. Other work has explored multimodal LLM frameworks that integrate sensor and voice data for continuous health monitoring, particularly in aging populations \cite{li2024understanding}.

Overall, existing literature highlights that: (1) LLMs are effective for extracting opioid-related and pain-related information from EHRs, and (2) they can model behavioral and contextual variables—such as diet or emotional tone—linked to opioid outcomes. However, relatively little is known about how LLMs perform when integrating multimodal data sources like wearable sensors and clinical records to forecast dynamic clinical phenomena such as pain spikes. Our current study addresses this unmet need by combining wearable and EHR data analyzed through multiple LLMs to enhance clinical insight into OUD and chronic pain comorbidity.

\section{Methods}
\textbf{Study design}
Twenty-five patients with CP and OUD who received MOUD at the APT Foundation were recruited for the two-week study. Following written completion of informed consent, the research assistant administered baseline measures (approximately 60 minutes). APT Foundation in New Haven, Connecticut, is a nonprofit organization that provides comprehensive addiction treatment including MOUD, counseling services, intensive outpatient and residential programs, and onsite psychiatric and primary medical care in an integrated model \cite{madden2018investigation}. Eligible patients were provided with a wearable device for two weeks and asked to complete a three-item paper and pencil pain measure. The research assistant (RA) met with eligible patients twice after baseline (end of weeks 1 and 2) to record data from the PEG-3 (see Table \ref{tab:schedule}) and to administer weekly assessments. The study was conducted in a single clinic, with care given to represent the demographic composition of the program. 

\textbf{Assessments}

\begin{table*}[t]
\centering
\resizebox{2.1\columnwidth}{!}{
\begin{tabular}{|c|c|}
\hline
Assessment name & Area of assessment \\
\hline
Perceived Stress Scale (PSS) \cite{cohen1983perceived} & Stress \\
\hline
General Self-Efficacy Scale (SES) \cite{sherer1994self} & Belief in one's competence to cope with stress \\
\hline
Substance Use Calendar & Substance use patterns \\
\hline
PEG-3 \cite{krebs2009development} & Daily measure of pain \\
\hline
Pittsburgh Sleep Quality Index (PSQI) \cite{smyth1999pittsburgh} & Sleep quality \\
\hline
Patient Health Questionnaire PHQ-9 \cite{choo2001phq} & Detection of depression \\
\hline
GAD-7 (General Anxiety Disorder-7) \cite{spitzer2006brief} & Detection of anxiety \\
\hline
Behavior and Symptom Identification Scale-24 (BASIS-24) \cite{cameron2007psychometric} & Mental health treatment outcomes \\
\hline
\end{tabular}
}
\caption{Overview of assessments}
\label{tab:overview}
\end{table*}

Assessments given to patients were selected by two experts who had published $>$10 peer-reviewed articles around OUD and CP, broadly defined. Experts independently developed lists of assessments based on study goals, clinical validity and ease of patient completion. Experts then compared the lists to result in the assessments indicated in Table \ref{tab:overview}. A third expert resolved disagreements in assessment selection. 

\textbf{Recruitment}
We collected data from 25 patients with chronic pain and opioid use disorder from the APT Foundation. Participants were recruited by study flyers and word-of-mouth. Inclusion criteria were as follows 1) 18 years or older; 2) Received MOUD at the APT Foundation; 3) Had chronic pain (greater than three months duration of pain occurring most days which limits life or work activities and/or leads to inability to work); 4) Understood English; 5) Able to provide informed consent; and 6) Owned a smartphone. The following were the exclusion criteria: 1) Had pending surgery or invasive pain management procedure; 2) Acutely psychotic, suicidal, or homicidal; 3) Had a pending or planned relocation or incarceration.

\textbf{Procedures}
Patients attended three study visits. During the first study visit, lasting approximately 60 minutes, prospective participants completed informed consent and baseline measures (see Assessment schedule in Table \ref{tab:schedule}). Participants were provided a wearable device (Fitbit Charge 5) to wear for two weeks and were asked to complete a three-item paper-and-pencil pain measure (PEG-3) daily. The wearable device continuously streamed a range of passively sensed patient data, such as heart rate variability, stress, and sleep. The research assistant met with eligible patients twice after baseline (end of weeks 1 and 2) to record data from the PEG-3 and to administer weekly assessments (Table \ref{tab:schedule}).

\begin{table}[t]
\centering
\resizebox{\columnwidth}{!}{
\begin{tabular}{|c|c|}
\hline
First meeting & Meetings at the end of Weeks 1 and 2 \\
\hline
Demographic information & Substance Use Calendar \\
\hline
PSS & SES \\
\hline
Substance Use Calendar & PEG-3 \\
\hline
SES & PSQI \\
\hline
PSQI & PHQ-9 \\
\hline
PHQ-9 & GAD-7 \\
\hline
GAD-7 &  --\\
\hline
BASIS-24 & --\\
\hline
\end{tabular}
}
\caption{Schedule of assessments}
\label{tab:schedule}
\end{table}

\textbf{Data sources}
\textit{Assessments}
All patient responses to assessments were noted and collected in a single data file. 
\textit{Wearable device: Fitbit Charge 5}
The Fitbit Charge 5 wristband \cite{fitbitcharge5}, is a popular smart wearable device widely utilized for fitness, activity, and wellness monitoring purposes. It gathers a range of information on physical and physiological aspects, such as heart activity rate, step count, floors climbed, calories burned, and various physical activities such as running and walking. Additionally, it tracks sleep quality, including metrics like light, deep, REM sleep, and overall sleep duration. Notably, the device also captures psychological indicators like stress, which is derived from physical signals, specifically heart rate. To establish connectivity between the wearable and the patient’s smartphone, the Fitbit app was installed and paired via Bluetooth. The app served as a conduit for collecting data from the wearable, which was subsequently transmitted to our data-stream-collecting servers and stored in password protected databases. All collected data was anonymized, and encrypted to ensure privacy and security. From the gathered signals, we computed daily summaries, providing insights into patients’ wellness and activity patterns.

\textbf{Features}
We used 180 candidate features based on data from the wearable device and assessments. We computed the daily mean, median, mode, minimum, and maximum for all features collected as a time series from the wearable device. Examples of features collected from the wearable device included sleep (duration for light, deep, REM sleep) and bedtime, daily step counts, daily floors climbed, and physical activity (duration of light, medium, heavy activity). 

\textbf{Predicting pain outcomes}
We sought to predict the probability of morning and evening pain spikes using data $\ge$a day before a spike. Pain spikes were defined as $>70$th percentile of self-reported pain. The use of a within-subject percentile threshold (e.g., $>70$th percentile) provided a personalized benchmark for identifying significant deviations from a person’s baseline. Percentile-based definitions of abnormal events or spikes have been used in other domains such as heart rate variability, stress monitoring, and mood disorders, where values exceeding the 70th–90th percentile are considered significant deviations worthy of clinical attention \cite{bartholomew2025review,o2016reference}. With only 25 patients, using a very high threshold (e.g., 90th percentile) would result in very few spike instances per patient, reducing model training effectiveness. The 70th percentile offered a pragmatic compromise that allowed us to model rare but recurrent pain spike events, while retaining sufficient data points for statistical inference and machine learning prediction. All features were used for prediction e.g., sleep, activity, PHQ-9, PEG-3. We computed all models by randomly splitting (75:15) the dataset into a training and a test dataset. We treated daily mean features as unique rows for model training. We used PerpetualBooster, a gradient boosting machine (GBM) algorithm which does not need hyperparameter optimization unlike other GBM algorithms. 

We also used several LLMs for providing insight around pain spikes. LLMs used: Claude Opus 4.5; Gemini 3 Pro; Gemini 2.5 Flash; Medgemma and Qwen3-7B Large. LLMs were selected based on their past uses to interpret health data, and the prompts were designed in collaboration with medical informaticians. Each row of the data was fed into the LLMs with the below prompts. All data was fed into the LLMs (wearable, surveys, EHR). We explored variations in results with different data inputs: Wearable only; EHR only (all survey data+patient demographics); Wearable+EHR. 

\begin{itemize}
    \item Prompt 1: I am providing you wearable device data from a patient with opioid use disorder and chronic pain, from a single day. I have also provided the headers for each feature. Based on these features, tell me a potential treatment plan.
    \item Prompt 2: I am providing you wearable device data from a patient with opioid use disorder and chronic pain, from a single day. I have also provided the headers for each feature. Based on these features, tell me if there is likely to be a morning pain spike ($>70$th percentile self-reported pain).
    \item Prompt 3: I am providing you wearable device data from a patient with opioid use disorder and chronic pain, from a single day. I have also provided the headers for each feature. Based on these features, tell me if there is likely to be an evening pain spike ($>70$th percentile self-reported pain).
    \item Prompt 4: I am providing you wearable device data from a patient with opioid use disorder and chronic pain, from a single day. I have also provided the headers for each feature. Based on these features, tell me what features are likely to be related to a morning pain spike ($>70$th percentile self-reported pain) the next day.
    \item Prompt 5: I am providing you wearable device data from a patient with opioid use disorder and chronic pain, from a single day. I have also provided the headers for each feature. Based on these features, tell me what features are likely to be related to a evening pain spike ($>70$th percentile self-reported pain) the next day.
\end{itemize}

\section{Results}
\begin{table}[]
    \centering
    \begin{tabular}{|c|c|}
    \hline
    Assessment   &  Mean (SD)\\ \hline
    PSS & 20.6 (6.19) \\ \hline
    SES & 30.00 (5.39)\\ \hline
    PEG-3 & 4.23 (1.39) \\ \hline 
     PSQI & 14.70 (3.61) \\ \hline 
    PHQ-9 & 10.92 (6.03) \\ \hline
    GAD-7 & 9.24 (6.42) \\ \hline
     BASIS-24  &  0.51 (0.10) \\ \hline
     
    \end{tabular}
        \caption{Assessments}
    \label{tab:assessment}
\end{table}

\begin{table}[]
    \centering
    \begin{tabular}{|c|c|}
        \hline
      Features   &  Mean (SD)\\ \hline
       Compliance  &  0.86 (0.23)\\ \hline
       Light sleep (mins) & 171.25 (174.28) \\ \hline
       Deep sleep (mins) & 28.61 (37.65)\\ \hline
       Total naps &  0.46 (0.71) \\ \hline
       Total time asleep (mins) & 361.48 (207.17)\\ \hline
       Sedentary activity (mins) & 890.92 (380.96)\\ \hline
       Light activity (mins) & 190.91 (121.99)\\ \hline 
       Moderate activity (mins) & 23.79 (42.91)\\ \hline
       High activity (mins) & 21.80 (43.05)\\ \hline
       Total daily step count &  5823.04 (6255.52) \\ \hline
    \end{tabular}
          \caption{Wearable features}
    \label{tab:wearable}
\end{table}

Patient mean age was 44.7 (SD=10.8) years. The study included 11 female and 14 male patients. Regarding race/ethnicity, we noted the following breakdown: Hispanic/Latino=4; white=17; Black=2, Other=2. Mean daily methadone dose was 102mg (SD=36.3). We provided means for the assessments in Table \ref{tab:assessment}. 60\% (N=15) of participants experienced pain spikes at least once during the study. Participants experienced an average of 3.2 (SD=1.6) pain spikes during the study. The average self-reported daily pain score on a scale of 10 was relatively high (M=5.8, SD=2.5), with similar findings on PEG-3 (M=12.7, SD=4.4). We also noted the probability of morning (M=0.26, SD=0.29) and evening pain (M=0.14, SD=0.20) spikes. SD of morning pain scores ranged from 0.97 to 1.83 (M=1.52). SD of evening pain scores ranged from 0.88 to 2.20 (M=1.61), indicating moderate fluctuations in self-reported pain levels across participants. There was a largely inconsistent relationship between morning and evening pain scores across participants, ranging from -0.35 to 0.08 (M=-0.05). PHQ-9 score was indicative of moderate depression (M=11, SD=6). GAD-7 score was similar (M=9.2, SD=6.4), perhaps reflecting moderate anxiety. PSS detailed moderate stress (M=20.6, SD=6.2), and PSQI indicated poor sleep quality (M=14.7, SD=3.6). 

Table \ref{tab:wearable} indicated that patients had a high compliance rate (M=0.86, SD=0.23). We defined compliance as the percentage of time that the device could successfully stream data to Fitbit servers, which we assumed was when the device was properly charged and worn. Compliance of 0 meant the device did not stream data throughout the study, and a compliance of 0.50 meant the device streamed data half the time. Mean compliance was 0.91 after excluding two outliers with 0.76 and 0.57 compliance. We sought to compare the mean patient characteristics with the Homekit 2020 Dataset \cite{merrill2023homekit2020}. The Homekit 2020 data is the largest public dataset of its kind, and patients were spread across 50 US states. Comparison of the data from the current study with Homekit 2020 allows for a better understanding of the distributions of characteristics in our data set. Patients in this study averaged 0.46 naps per day (see Table \ref{tab:wearable}), compared to the Homekit data of 0.14 naps per day. However, our patients slept fewer total minutes, with an average of 361 minutes compared to 460 minutes in the Homekit data. Our patients recorded more daily sedentary time, averaging 890 minutes per day compared to the Homekit baseline of 688 minutes, and engaged in less light physical activity, averaging 190 minutes per day versus 239 minutes.
\textbf{Predicting pain outcomes} 

\begin{table}[]
    \centering
    \begin{tabular}{|c|c|c|}
    \hline
    {Outcome} & {Time interval} & {Accuracy} \\ \hline
    \multirow{5}{*}{Morning spike} & t+1 & 0.68 \\ \cline{2-3}
    & t+2 & 0.73 \\ \cline{2-3}
    & t+3 & 0.77 \\ \cline{2-3}
    & t+4 & 0.79 \\ \cline{2-3}
    & t+5 & 0.81 \\ \hline
    \multirow{5}{*}{Evening spike} & t+1 & 0.86 \\ \cline{2-3}
    & t+2 & 0.90 \\ \cline{2-3}
    & t+3 & 0.94 \\ \cline{2-3}
    & t+4 & 0.90 \\ \cline{2-3}
    & t+5 & 0.81 \\ \hline
    \end{tabular}
    \caption{Model accuracy}
    \label{tab:performance}
\end{table}

Table \ref{tab:performance} indicated model accuracy for the probability of a spike in morning and evening pain. Broadly, we noted relatively high accuracy ($>0.7$) in the prediction of morning and evening pain spikes for one (t+1) to five days (t+5) in the future. These results may indicate the utility of using wearable data to predict pain spikes a few days in the future. Combining wearable data with health record data may improve model performance, and aid in the management of pain spikes in the population of CP/OUD patients. We conducted SHAP analysis for the prediction of the next day's morning and evening pain spikes. Morning step count, evening step count, and time spent asleep had a high impact on predicting a morning pain spike. Similarly, light activity, time asleep, and morning step count had a large impact on predicting an evening pain spike. 

\newcolumntype{L}{>{\raggedright\arraybackslash}m{3cm}}

\begin{table*}[t]
\centering
\resizebox{2.0\columnwidth}{!}{
\begin{tabular}{|L|L|L|L|L|}
\hline
\multicolumn{1}{|c|}{Claude-Opus 4.5} &
\multicolumn{1}{c|}{Gemini 2.5-Flash} &
\multicolumn{1}{c|}{Gemini 3-Pro} &
\multicolumn{1}{c|}{MedGemma} &
\multicolumn{1}{c|}{Qwen3-7B} \\ \hline

Next-day morning pain spikes are commonly associated with activity imbalance and prolonged inactivity. Extremely high sedentary time, very low moderate and vigorous activity, and a low step count suggest pain avoidance and increased pain sensitivity.
&
Next-day morning pain spikes are predicted by high previous-evening pain, worsening daily pain, and poor overnight recovery, including fragmented sleep and psychological stress.
&
Morning pain spikes usually follow worsening pain and non-restorative sleep, with elevated evening pain and activity extremes as key predictors.
&
High evening pain (8) and sustained pain into the next morning (8) indicate unresolved pain and a likely next-day flare.
&
A morning pain spike is unlikely; elevated pain reflects persistence rather than a sharp deviation from baseline. \\ \hline

Next-day evening pain spikes are often associated with poor overnight recovery. Extremely short sleep duration and insufficient restorative sleep increase vulnerability.
&
Evening pain spikes result from accumulated strain, high baseline pain, poor sleep, activity extremes, and psychosocial stressors.
&
Unresolved pain, poor sleep quality, and activity extremes increase next-day evening flare risk.
&
Pain persistence and activity imbalance are strongest predictors, with high evening pain most indicative.
&
High risk of evening pain spike driven by a large pain increase from morning (4) to evening (8) and sustained pain. \\ \hline
\end{tabular}
}
\caption{LLM insights}
\label{tab:LLM}
\end{table*}

Table \ref{tab:LLM} presents a comparison of large language model (LLM) outputs for predicting next-day morning and evening pain spikes using wearable, pain, and mental health features. Clear differences emerged in how models reasoned about patient-specific data, spike definitions, and prediction confidence.
Claude-Opus 4.5 consistently produced the most clinically interpretable outputs. The model integrated observed wearable features (e.g., sedentary time, activity levels), pain trajectories, and baseline comparisons to distinguish persistent pain from true pain spikes. Notably, Claude explicitly concluded that a morning pain spike was unlikely when current morning pain did not exceed the patient’s baseline, while simultaneously identifying a high risk for an evening pain spike driven by large increases from morning to evening pain and sustained high pain into the next day.
Qwen3-7B also demonstrated strong patient-grounded reasoning. The model explicitly compared observed pain values to the patient’s mean pain level and correctly ruled out a morning pain spike, attributing elevated next-morning pain to persistence from the prior evening rather than a distinct spike. Qwen3-7B further identified high risk for an evening pain spike, citing large within-day pain increases, activity imbalance, psychological distress, and opioid use as contributing factors.
In contrast, Gemini 2.5-Flash and Gemini 3-Pro primarily provided generalized descriptions of known predictors of pain spikes, such as high evening pain, poor sleep quality, and extremes in physical activity. While these models produced coherent summaries of relevant risk factors, they did not anchor predictions to the patient’s observed values or provide explicit determinations of spike likelihood.
MedGemma generated predictive statements but showed weaker alignment with the defined spike criteria. The model did not consistently distinguish between persistent pain and threshold-based pain spikes and provided less explicit justification using patient-specific baseline comparisons.

Overall, the findings indicate substantial variability in LLM performance. Models that explicitly referenced patient-specific values and baseline thresholds (Claude-Opus 4.5 and Qwen3-7B) produced clearer and more clinically actionable pain spike assessments than models that remained largely descriptive or inconsistently applied spike definitions.

\section{Discussion}
\textbf{Implications of Findings} 
Our findings suggested that machine learning models can achieve relatively high accuracy ($>0.7$) in predicting pain spikes. SHAP analysis highlighted several key features that contributed to pain spike prediction, including morning and evening step counts, sleep duration, and light activity levels. Real-time monitoring through wearable devices, combined with advanced AI models, could facilitate early detection of pain spikes and support personalized interventions \cite{goyal2023predicting,shimgekar2025agentic,shimgekar2025nimblelabs}. This approach may help mitigate the risk of opioid relapse, improve adherence to MOUD, and enhance the integration of CP and OUD care. LLM analysis revealed sub-par model performance across data contexts, highlighting the need to develop LLMs which can provide actionable insights in the OUD/CP context \cite{van2023fatherhood,chen2022categorizing,goyal2023misinformative}.

\textbf{Limitations and future work}
While our sample size was adequate for a proof-of-concept study, it limits the generalizability of our findings to larger and more diverse populations. Future research should recruit larger cohorts with greater diversity in demographics, clinical characteristics, and smartphone ownership to improve model robustness and external validity. Our wearable data was limited to the Fitbit device. Future studies should explore multimodal approaches that combine wearable data with other sensor types (e.g., voice or environmental sensors) to capture a broader range of pain-related features. The LLM models showed varying levels of performance and interpretability \cite{goyal2024using,ligo2024user,vassef2025one}. Further research should refine LLMs to better handle incomplete or multimodal data while addressing concerns related to data privacy, ethical decision-making, and bias \cite{goyal2023chatgpt}. Developing context-aware LLMs that can offer meaningful insights even with partial data could improve their application in clinical practice. The extent to which the operational definition of pain spike used in this study (greater than 70th percentile of self-reported pain) is clinically meaningful is currently unclear and should be assessed in future studies.

\bibliography{chil-sample}
\end{document}